\documentclass[10pt,twocolumn,letterpaper]{article}

\usepackage{cvpr}              % To produce the CAMERA-READY version

\usepackage[accsupp]{axessibility}
\usepackage{graphicx}
\usepackage{amsmath}
\usepackage{amssymb}
\usepackage{booktabs}
\usepackage{multirow}
\usepackage{makecell}
\usepackage{adjustbox}

\usepackage[pagebackref,breaklinks,colorlinks]{hyperref}

\usepackage[capitalize]{cleveref}
\crefname{section}{Sec.}{Secs.}
\Crefname{section}{Section}{Sections}
\Crefname{table}{Table}{Tables}
\crefname{table}{Tab.}{Tabs.}

\def\cvprPaperID{3199} % *** Enter the CVPR Paper ID here
\def\confName{CVPR}
\def\confYear{2023}

\begin{document}

%%%%%%%%% TITLE - PLEASE UPDATE
\title{Point2Pix: Photo-Realistic Point Cloud Rendering via Neural Radiance Fields}

\author{
Tao Hu$^{1}$ \quad Xiaogang Xu$^{1}$\footnotemark[1] \quad Shu Liu$^{2}$ \quad Jiaya Jia$^{1,2}$\\
$^1$ The Chinese University of Hong Kong \quad $^2$ SmartMore\\
{\tt \small \{taohu, xgxu, leojia\}@cse.cuhk.edu.hk, liushuhust@gmail.com}
%%First Author\\
%%Institution1\\
%%Institution1 address\\
%%{\tt\small firstauthor@i1.org}
% For a paper whose authors are all at the same institution,
% omit the following lines up until the closing ``}''.
% Additional authors and addresses can be added with ``\and'',
% just like the second author.
% To save space, use either the email address or home page, not both
%%\and
%%Second Author\\
%%Institution2\\
%%First line of institution2 address\\
%%{\tt\small secondauthor@i2.org}
}
\maketitle

\renewcommand{\thefootnote}{\fnsymbol{footnote}} 
\footnotetext[1]{Corresponding author.}

\begin{abstract}
	Synthesizing photo-realistic images from a point cloud is challenging because of the sparsity of point cloud representation. Recent Neural Radiance Fields and extensions are proposed to synthesize realistic images from 2D input. In this paper, we present Point2Pix as a novel point renderer to link the 3D sparse point clouds with 2D dense image pixels. Taking advantage of the point cloud 3D prior and NeRF rendering pipeline, our method can synthesize high-quality images from colored point clouds, generally for novel indoor scenes. To improve the efficiency of ray sampling, we propose point-guided sampling, which focuses on valid samples.  Also, we present Point Encoding to build Multi-scale Radiance Fields that provide discriminative 3D point features. Finally, we propose Fusion Encoding to efficiently synthesize high-quality images. Extensive experiments on the ScanNet and ArkitScenes datasets demonstrate the effectiveness and generalization.
\end{abstract}

\section{Introduction}
Point cloud rendering aims to synthesize images from point clouds at given camera parameters, which has been frequently utilized in 3D visualization, navigation, and augmented reality. There are many advantages to point cloud representation, such as flexible shape and general 3D prior. However, 
since point clouds are generally produced by 3D scanners (RGBD  or LiDAR) \cite{KITTI,Arkitscenes,ScanNet} or by Multi-View Stereo (MVS) from images \cite{MVS,openmvs2020,MVSNet},  the points are usually sparsely distributed in 3D scenes.  Although traditional graphics-based renderers \cite{PCL,Open3D,Pytorch3d,OpenGL} can render point clouds to images without training or finetuning, the quality is not satisfying with hole artifacts and missing details \cite{NPCR}.

Recently, Neural Radiance Fields (NeRF) \cite{NeRF}  were proposed for 3D representation and high-fidelity novel view synthesis. It employs an implicit function to directly map each point's spatial information (location and direction) to attributes (color and density). However, most NeRF-based methods \cite{NSVF,PlenOctree,NeRF++,NeRF--} are scene-specific, thus taking much time to train from scratch for novel scenes in abundant multi-view images, which limits the practical applications. 

In this work, we bridge the gap between point clouds and NeRF, thus proposing a novel point cloud renderer, called Point2Pix, to synthesize photo-realistic images from colored point clouds. Compared with most NeRF-based methods \cite{NeRF,NSVF,PlenOctree,Plenoxels}, ours does not necessarily require multi-view images or fine-tuning procedures for indoor scenes. 

First, point clouds are treated as underlying anchors of NeRF. The training process of NeRF is to learn 3D point attributes from given locations. Because there is no mapping ground truth for multi-view images, NeRF-based methods \cite{NeRF,NSVF} indirectly train their networks with a pixel reconstruction loss. Note that point clouds are exactly made up of points with location and attributes, thus can provide training pairs for the mapping function to conduct supervised learning and improve performance. 

Then, point clouds can also improve the efficiency of ray sampling. NeRF-based methods \cite{NeRF,NSVF,PlenOctree} learn the 3D shape and structure from multi-view images, which do not involve geometric prior in novel scenes. Thus, dense uniform sampling \cite{MVSNeRF,IBRNet} or coarse-to-fine sampling \cite{NeRF,PlenOctree} were used to synthesize high-quality images. These strategies are inefficient because most of the locations in 3D scenes are empty \cite{NSVF,DO-NeRF}. Since point clouds represent a relatively fine shape of 3D scenes, the area around existing points deserves more attention. Based on this observation, we propose a point-guided sampling strategy to mainly focus on the local area of points in the point cloud. It can significantly reduce the number of required samples while maintaining decent synthesis accuracy.  

Further, point-based networks can provide 3D features prior to subsequent applications, general for novel scenes. Although many methods \cite{PN,PointNet++,SparseConvNet,ME} have been proposed for various point cloud understanding tasks, they are usually designed for existing points.  In this work, we propose Multi-scale Radiance Fields, including Point Encoder and MLP networks, to extract multi-scale features for any location in the scene. These 3D point features are discriminative and general, ensuring finetuning-free rendering. Also, inspired by recent NeRF-based image generators \cite{StyleNeRF,CIPS-3D,HeadNeRF}, we render the 3D point features as multi-scale feature maps. Our fusion decoder gradually synthesizes high-resolution images. It can not only fill the possible holes but also improve the quality of rendered images. Our main contributions are summarized as follows.
\begin{itemize}
	\item We propose Point2Pix to link point clouds with image space, which renders point clouds into photo-realistic images.   
	\item We present an efficient ray sampling strategy and fusion decoder to greatly decrease the number of samples in each ray and the total number of rays, thus accelerating the rendering process.  
	\item We propose Multi-scale Radiance Fields, which extract discriminative 3D prior for arbitrary 3D locations.
 %in 3D scenes.  
	%	\item Our Point2Pix can also be applied to downstream tasks, such as point cloud upsampling and semantic-to-image translation. 
	\item Extensive experiments and ablation studies on indoor datasets demonstrate the effectiveness and generalization of the proposed method.
\end{itemize}

\section{Related Work}
In this section, we briefly review the related works, including various point renderers, point-based networks in extracting 3D point features, and NeRF-based synthesis.
%in different scenes.

\subsection{Point-based Rendering}
Traditional Point Rendering \cite{Pytorch3d,PCL,Open3D} is based on computer graphics, which generates images from point clouds by simulating the physical process of imaging, considering geometry \cite{Photo_tourism}, material \cite{Material}, BRDF \cite{BRDF}, and lighting \cite{Lighting}. The rendering pipeline is general for arbitrary scenes. But it cannot fill missing points and thus generate vacant pixels. 

In the deep learning era, neural-based point renderers \cite{NPBG,NPCR,ADOP} made great progress in generating images from point clouds. They first extract multi-scale projected features from point clouds and then use a decoder to generate images. NPBG \cite{NPBG} augments each point by a neural descriptor to encode local geometry and appearance. NPCR \cite{NPCR} represents point features by 3D Convolution Neural Network (CNN), and converts the 3D representation to multi-plane images. Different from these methods, our Point2Pix combines a more discriminative point encoder with the renderer pipeline of NeRF, thus achieving better performance.

\subsection{Point-based Networks}
Point-based networks have been developed for many years \cite{PN,PointNet++,Icm-3d,Sparse_3D_CNN,SparseConvNet,ME}. For general point understanding, PointNet \cite{PN} utilizes point-wise Multi-Layer Perception (MLP) and Pooing to extract features for 3D classification and semantic segmentation, while not capturing local structures. PointNet++ \cite{PointNet++} introduces hierarchical feature learning to encode local point features. Although 3D CNN can also deal with point cloud data after voxelization, the maximal resolution of 3D volumes is low because of huge memory consumption and slow computation. Thus, sparse 3D CNNs \cite{Sparse_3D_CNN,SPLATNet,Monte_Carlo_convolution,ME} draw more attention. SparseConvNet \cite{SparseConvNet} differs from previous sparse convolutional networks since it uses a rectangular grid, instead of dilating the observation. In this paper, we adopt a more efficient sparse 3D CNN -- MinkowskiEngine \cite{ME} -- as the basic point encoder to extract 3D prior from point clouds. 

\subsection{NeRF-based Synthesis}
NeRF \cite{NeRF} well balances neural network and physical rendering, thus achieving state-of-the-art performance in novel view synthesis. To handle dynamic scenes where objects move, a deformable function is learned to align different frames \cite{nerfies,NSFF}. For human reconstruction and synthesis, many methods, such as Neural-Body\cite{Neural-Body}, Neural-Actor \cite{Neural-Actor}, and Anim-NeRF \cite{Anim-NeRF}, introduce the parameterized human model SMPL \cite{SMPL} as a strong 3D prior and achieve impressive performance. To generate high-resolution images, methods of \cite{GIRAFFE,HeadNeRF,StyleNeRF,CIPS-3D} first render low-resolution feature maps instead of images, then upsample features to the final images via 2D CNN. NeRF's input is only multi-view images and camera parameters, when combined with 3D prior, such as depth and point cloud, the performance can be further improved \cite{PointNeRF,Neural_RGBD,DO-NeRF,DS-NeRF}. Our model also combines deep learning with physical rendering, while taking more advantage of point clouds as 3D priors to render decent-quality images in general indoor scenes.

\section{Our Approach}
For a point cloud $\textbf{P}= \cup_{k=1}^K  \{ \textbf{p}_k, \textbf{c}_k \} $ with $K$ points, $\textbf{p}_k = (x_k, y_k, z_k)\in \mathbb{R}^{3}$ and corresponding colors $ \textbf{c}_k = (r_k,g_k,b_k)\in \mathbb{R}^{3}$, our goal is to synthesize a high-fidelity image $\hat{I}$ at the given camera parameter $\textbf{V}$ via our proposed renderer (Point2Pix) $\mathcal{R}$, formulated as
\begin{equation}
	I = \mathcal{R}(\textbf{P}, V),
	\label{eq:point_rendering}
\end{equation}
where $V$ is represented by $H\times W$ rays $\textbf{\textbf{r}}$. Each ray starts from camera center $\textbf{o}$ in pixel direction $\textbf{d}$.

\begin{figure*}[t]
	\centering
	\includegraphics[width=0.99 \linewidth]{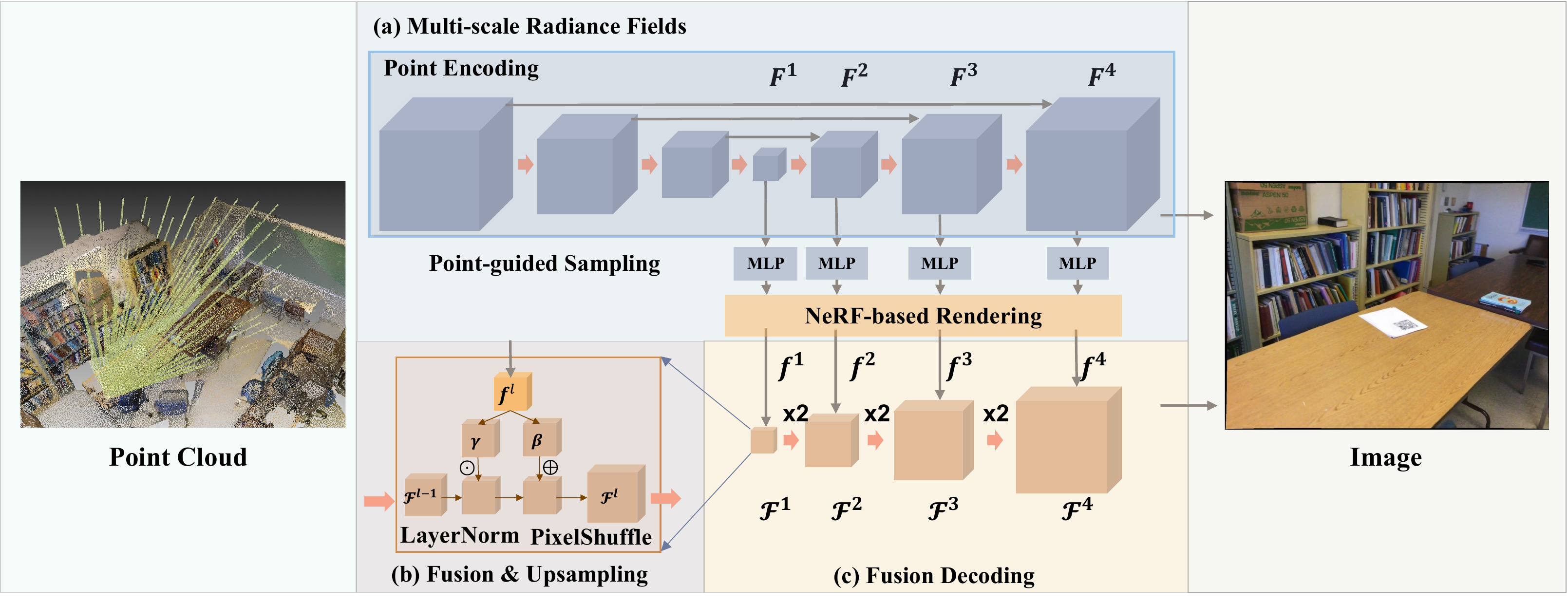}
	\caption{Overview of our proposed Point2Pix. (a) Multi-scale Radiance Fields. For an input point cloud, we first extract multiple 3D feature volumes in four scales. Next, for any queried point, we first linearly interpolate the coarse features from these feature volumes and infer the final features through MLP networks. (b) Fusion and Upsampling. Four 2D feature maps are respectively rendered through NeRF. They are fused with the previous 2D CNN output and then are upsampled by 2 times. (c) Fusion Decoding. We finally design a neural renderer to gradually synthesize target images from the projected feature maps.}
	\label{fig:overview}
\end{figure*}

To begin with, we introduce the background knowledge of NeRF in Sec. \ref{sec:nerf}. Then, we propose an efficient point-based ray sampling strategy in Sec. \ref{sec:point_sampling}. We also build a network to extract multi-scale 3D prior from point clouds in Sec. \ref{sec:multi_scale_radiance_fields}. Finally, we show how to combine the point feature with NeRF to render target images in Sec. \ref{sec:fusion}. The overview of our framework can be seen in Fig.~\ref{fig:overview}.

\subsection{Preliminary}
\label{sec:nerf}
Given multi-view camera-calibrated images of a scene, NeRF \cite{NeRF} synthesizes high-quality novel view images.  It mainly consists of ray sampling, implicit function, and volume rendering. 

\noindent\textbf{Ray Sampling.}
Starting from the camera center $\textbf{o}$, ray sampling is the process that obtains a series of positions $\textbf{x}_i$ along ray $\textbf{r}$ with direction $\textbf{d}$ as
\begin{equation}
	\textbf{x}_i = \textbf{o} + z_i \cdot \textbf{d}, \quad i = 1, 2, ..., N,
\end{equation}
where $z_i$ is the sampling depth and $N$ is the number of samples on each ray.

\noindent\textbf{Implicit Function}.
An implicit function $f_\theta$ is trained as a mapping from each queried location $\textbf{x}_i$ and direction $\textbf{d}$ to corresponding colors $\textbf{c}_i=(r_i,g_i,b_i)$ and density $\sigma_i$, as
%%formulated as
\begin{equation}
	(\textbf{c}_i, \sigma_i) = f_\theta(\textbf{x}_i, \textbf{d}),
	\label{eq:implicit_function}
\end{equation}
where $f_\theta$ is an MLP network, and $\theta$ is its parameter.

\noindent\textbf{Volume Rendering}.
Each ray (or pixel) color $\hat{\textbf{c}}$ is calculated via volume rendering \cite{volume_rendering} as
\begin{equation}
\begin{aligned}
	\label{eq:volume_rendering}
	\hat{\textbf{c}} &= \sum_{i=1}^{N} T_i \alpha_i \textbf{c}_i,\\ 
	\alpha_i &= 1 -\textbf{ exp}(-\sigma_i \delta_i),\\ 
	T_i &= \textbf{exp}(-\sum_{j=1}^{i-1}\sigma_j\delta_j),
\end{aligned}
\end{equation}
where $\delta_j$ is the distance between neighbor samples along the ray $\textbf{r}$.

\subsection{Point-guided Sampling}
\label{sec:point_sampling}
According to Eq. \eqref{eq:volume_rendering}, increasing the number of samples $N$ along each ray can generate more realistic results \cite{NeRF}. However, the required computing resources and running time will also linearly grow. Our model is based on a point cloud with a relatively fine shape prior. Thus, we propose point-guided sampling to achieve more efficient ray sampling through the guidance of the point cloud. 

\begin{figure}[t]
	\centering
	\includegraphics[width=0.5\linewidth]{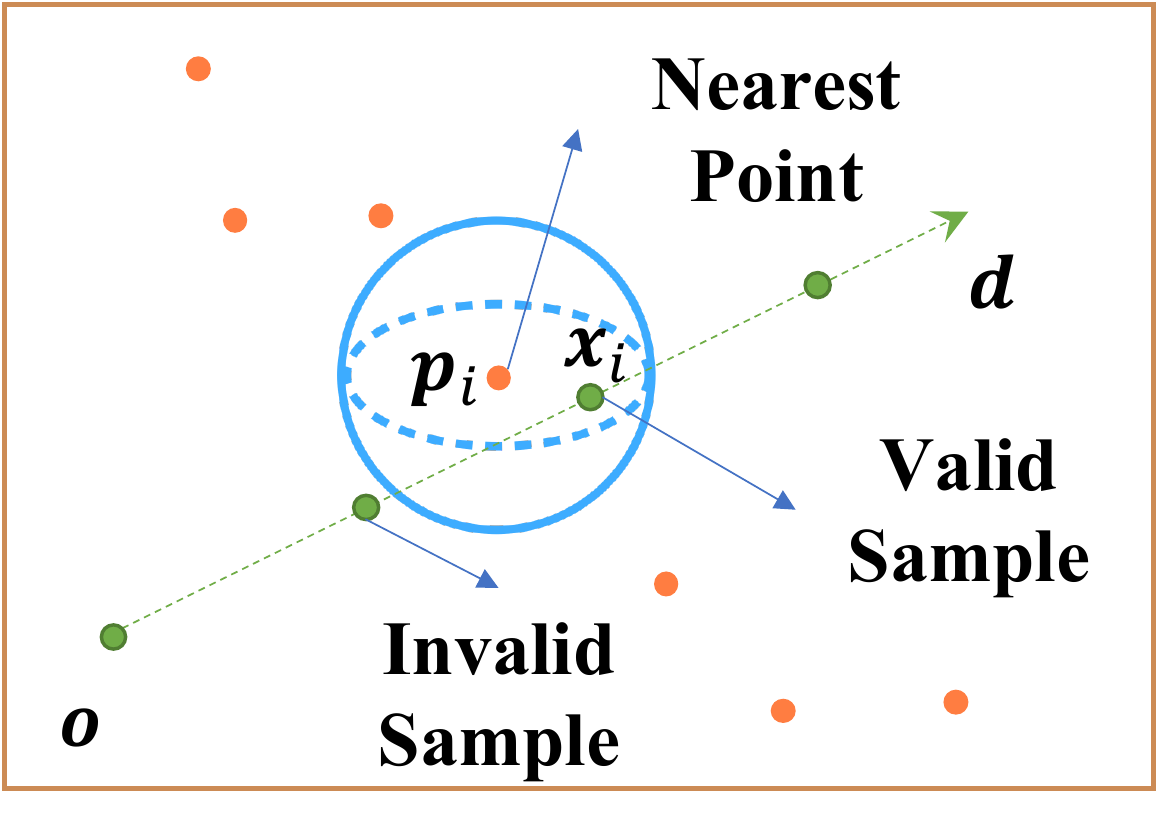}
	\caption{The proposed point-guided sampling. For any queried point $\textbf{x}_i$,  we find its nearest point $\textbf{p}_i$ in the point cloud. If $\textbf{x}_i$ is located in the ball area (with radius $r$) of $\textbf{p}_i$, it is a valid sample. Invalid samples are omitted to improve sampling efficiency. }
	\label{fig:sampling}
\end{figure}

For any queried point $\textbf{x}_i$, we first find the nearest neighbour point $\textbf{p}_i$, then check whether $\textbf{x}_i$ is located in $\textbf{p}_i$'s ball area with radius $r$ or not, as
\begin{equation}
	\parallel \textbf{p}_i - \textbf{x}_i \parallel_2 \leq r.
	\label{eq:sampling}
\end{equation}
If the above condition is satisfied, we treat the queried point $\textbf{x}_i$ as a valid sample and obtain the point feature in Sec. \ref{sec:multi_scale_radiance_fields}. If there is no valid sample along one ray, we adopt uniform sampling from default near to far depth. As illustrated in Fig. \ref{fig:sampling}. Compared with previous uniform and coarse-to-fine sampling \cite{NeRF,MVSNeRF,IBRNet}, our sampling strategy reduces computation and memory costs.

\subsection{Multi-scale Radiance Fields}
\label{sec:multi_scale_radiance_fields}
We extract discriminative 3D point and ray features via constructing Multi-scale Radiance Fields, including Point Encoding and NeRF-based Feature Rendering.

\noindent\textbf{Point Encoding}.
Point Encoding is to output a discriminative 3D point feature for each valid sample $\textbf{x}_i$. We adopt 3D sparse UNet from the Minkowski Engine (ME) \cite{ME} as the backbone of our point encoder. ME is an auto-differentiation library to build a 3D CNN for sparse tensors, that are converted from point clouds. As illustrated in Fig. \ref{fig:overview}, our point encoder extracts multiple 3D feature volumes from the raw point cloud in $L$ different scales. We select $F_l$ at scale $l$ to construct multi-scale radiance fields.  
%Inputting all points to the Point Encoder in the point cloud will consume a lot of memory according to Tab.  \ref{tab:memory}. Therefore, we only input points within a camera view $\theta$ to the encoder, as represented in Eq. \ref{eq:part_select}. 
%\begin{equation}
	%	\arccos_\textbf{x}(\textbf{x} - \textbf{o}, \textbf{x}_c - \textbf{o}) \ge \theta.
	%	\label{eq:part_select} 
	%\end{equation}
%where $\textbf{x}_c$ is the image's center point in the world coordinate, and $\theta$ should be greater than the camera's FOV angle. As shown in Tab. \ref{tab:camera_view}, the point outside of camera's Fileds Of View (FOV) only contributes little the final rendering results. 

For each valid sample $\textbf{x}_i$ at each scale $l$, we query feature in $F_l$ to obtain the interpolated feature $F_i^l$ and employ an implicit function $\Phi_l$ to infer density $\sigma_i^l$ and final point feature $f_i^l$ for $\textbf{x}_i$ as
\begin{equation}
	(\sigma_i^l, \textbf{f}_i^l) = \Phi_l(\textbf{F}_i^l) = \Phi_l(\textbf{F}^l[\textbf{x}_i]).
	\label{eq:feature_function}
\end{equation}

\noindent\textbf{NeRF-based Feature Rendering}.
Then we render the queried 3D point features to 2D feature maps and generates images at different scales. 
At each feature scale $l$, we aggregate density $\sigma_i^l$ and  feature $\textbf{f}_i^l$ to generate 2D feature map $\textbf{f}^l$ by volume rendering of NeRF \cite{NeRF} as
\begin{equation}
\begin{aligned}
	\textbf{f}^l &= \sum_{i=1}^{N} \textbf{w}_i^l \textbf{f}_i ^l, \\
	% \textbf{c}^l &= \sum_{i=1}^{N} \textbf{w}_i^l \textbf{c}_i ^l,  \\
	\textbf{w}_i^l &= \textbf{exp}(-\sum_{j=1}^{i-1}\sigma_j^l\delta_j) (1 - \textbf{exp}(-\sigma_i^l \delta_i)).
\end{aligned}
\label{eq:feature_rendering}
\end{equation}
So far, we obtained $L$ rendering feature maps $\{\textbf{f}^l\} \in \mathbb{R}^{C_l\times H_l\times W_l}$, where $H_l$ and $W_l$ respectively represent the feature height and width, %$\textbf{f}^l$, 
and $C_l$ is the number of channels.

\subsection{Fusion Decoding}
\label{sec:fusion}
Although we propose an efficient ray sampling strategy to reduce memory consumption, it still requires more than 20 GB  GPU memory to render target images with the size of 480 $\times$ 640, as shown in Tab. \ref{tab:multi_scale}. In addition, there are still many holes to be filled in the 2D-rendered image space. To address these issues, we design a Fusion Decoder as a neural renderer that synthesizes final images from these rendered feature maps $\textbf{f}^l, l\in [1, L]$ by conditional convolution and upsampling modules. 

\noindent\textbf{Fusion}.
Our conditional convolution is to fuse the previous layer's feature $\mathcal{F}^{l-1}$ with the rendered features $\textbf{f}^l$, which treats the rendered feature at each scale $l$ as the conditional input. This module is inspired by SPADE \cite{SPADE}, while we use Layer Normalization \cite{LN}. Specifically, as shown in Eq. \eqref{eq:fusion}, for the rendered feature map $\textbf{f}^l$, we calculate the conditional parameters, including scale $\gamma$ and bias $\beta$, by a $\textit{Conv2D}$ module. Then for feature $\mathcal{F}^{l-1}$ from the previous stage, we normalize it by Layer Normalization and scale it by $\gamma$. Finally, the fused feature $\mathcal{F}^l$ is obtained by adding the bias $\beta$ as
\begin{align}
	(\gamma, \beta) &=   \textbf{Conv2D}(\textbf{f}^l), \\ \nonumber
	\mathcal{F}^{l-1} &= \gamma \cdot \textbf{LayerNorm}(\mathcal{F}^{l-1}) + \beta.
	\label{eq:fusion}
\end{align}

\noindent\textbf{Upsampling}. We adopt PixelShuffle \cite{PixelShuffle} as our upsampling modules that upsample the fused feature $\mathcal{F}^{l-1}$ by $2$ times at each stage, instead of using bilinear or nearest interpolation. PixelShuffle \cite{PixelShuffle} is frequently adopted in the super-resolution task, which utilizes a convolution layer to extend the channel size and reshape them into the spatial size, as
\begin{align}
	\mathcal{F}^{l}&= \textbf{Pixelshuffle}(\textbf{Conv2D}(\mathcal{F}^{l-1}), 2).
\end{align}

\noindent\textbf{ToRGB}. Finally, we introduce the decoder of the present large-scale generator, like \cite{latent-diffusion}, as a post-process to generate the final rendering image $\hat{I}$ for the whole point cloud renderer.

\begin{table*}[t]
	\begin{adjustbox}{width=0.8\linewidth,center}
		\begin{tabular}{c|ccc|ccc}
			\hline
			Dataset & \multicolumn{3}{c|}{ScanNet \cite{ScanNet}} & \multicolumn{3}{c}{ARKitScenes \cite{Arkitscenes}} \\ \hline
			Metrics & \quad PSNR $\uparrow$ & \quad SSIM $\uparrow$ & \quad LPIPS $\downarrow$ & \quad PSNR$\uparrow$  & \quad SSIM$\uparrow$ & \quad LPIPS$\downarrow$ \\ \hline
			Pytorch3D \cite{Pytorch3d} & 13.62 & 0.528 & 0.779  & 15.21 & 0.581 & 0.756  \\
			Pix2PixHD \cite{Pix2PixHD} & 15.59 & 0.601 & 0.611 & 15.94 &  0.636  & 0.605  \\
			NPCR \cite{NPCR} & 16.22  & 0.659 & 0.574  & 16.84  &0.661  & 0.518  \\
			NPBG++ \cite{NPBG} & 16.81 & 0.671 & 0.585 & 17.23 & 0.692 & 0.511 \\
			ADOP \cite{ADOP} & 16.83  & \textbf{0.699} & 0.577  & 17.32  &0.707  & \textbf{0.495}  \\
            Point-NeRF \cite{PointNeRF} & \textbf{17.53} & 0.685 & \textbf{0.517} & \textbf{17.61} & \textbf{0.715} & 0.508 \\
			\hline
			\textbf{Point2Pix (Ours)} & \textbf{18.47} & \textbf{0.723} & \textbf{0.484 }& \textbf{18.84}  &\textbf{0.734} &\textbf{ 0.471 }\\ \hline
		\end{tabular}
	\end{adjustbox}
	%%\vspace{0.1 in}
	\caption{Comparing our method with different point renderers on the ScanNet \cite{ScanNet} and ARkitScenes \cite{Arkitscenes} datasets. There is no finetuning process in this experiment, which demonstrates the generalization in novel scenes. }
	%\vspace{-0.1 in}
	\label{tab:render_comparison}
\end{table*}

\subsection{Loss Function}
For the NeRF-based rendering images and neural rendering images, their optimization goals are the ground-truth images with target camera parameters. We employ point cloud loss, NeRF rendering loss, neural rendering loss, and perceptual loss to train the parameters of the proposed point encoder and fusion decoder as
% \begin{equation}
% 	\mathcal{L} = \lambda_{pc} \mathcal{L}_{pc} + \lambda_{nerf} \mathcal{L}_{nerf} +  \lambda_{nr} \mathcal{L}_{nr} + \lambda_{per} \mathcal{L}_{per}, 
% \end{equation}
\begin{equation}
	\mathcal{L} = \lambda_{pc} \mathcal{L}_{pc} +  \lambda_{nr} \mathcal{L}_{nr} + \lambda_{per} \mathcal{L}_{per}, 
\end{equation}
where $\lambda_{pc}$,  $\lambda_{nr}$, and  $\lambda_{per}$ respectively control weights of these losses. 

\noindent\textbf{Point Cloud Loss}. All points $\textbf{p}_k$ in raw point clouds provide groud-truth mapping from locations $\textbf{x}_k$ to densities $\hat{\sigma}_k$ and colors $\hat{\textbf{c}}_k$. Denoting the queried 3D densities and colors from Point2pix in point $\textbf{p}_k$  as $\textbf{c}_k$ and  $\sigma_k$, the point cloud loss can be represented as
\begin{equation}
	\mathcal{L}_{pc} = \sum_{k=1}^{K} ( \parallel  \hat{\textbf{c}}_k - \textbf{c}_k \parallel^2 +  \frac{1}{D} \max(0, D-\sigma_k)).
	\label{eq:loss_pc}
\end{equation}
We encourage the predicted densities at $\textbf{p}_k$ to be greater than a threshold $D$. 

% \noindent\textbf{NeRF Rendering Loss}. $\mathcal{L}_{nerf}$ indicates the sum of square error between the NeRF rendering ray colors $\textbf{c}^l$ and ground-truth ray colors  $\hat{\textbf{c}}^l$ at each scale $l$ as
% \begin{equation}
% 	\mathcal{L}_{nerf} = \sum_{l=1}^L \parallel \hat{\textbf{c}}^l - \textbf{c}^l \parallel^2.
% \end{equation}

\noindent\textbf{Neural Rendering Loss}. $\mathcal{L}_{nr}$ is the MSE between rendered images $I$ from fusion decoder and ground truths $\hat{I}$ as
%%square error between the rendered images $I$ from fusion decoder and the ground-truth images $\hat{I}$ as
\begin{equation}
	\mathcal{L}_{nr} = \parallel \hat{I} - I \parallel^2_2.
\end{equation}

\noindent\textbf{Perceptual Loss}. $\mathcal{L}_{per}$ is a frequently used loss in image synthesis, which improves the realism of generation, as
%%which improves the realism of generated images as
\begin{equation}
	\mathcal{L}_{per} =  \sum_{l=1}^{L} \parallel \phi(\hat{I}^l)- \phi(I^l )\parallel_1,
\end{equation}
where $\phi(\cdot)$ means extracting VGG features.

\section{Experiments}
In this section, we conduct experiments to demonstrate the effectiveness of our proposed method. First, we introduce the indoor datasets and evaluation metrics. Then, we quantitatively and qualitatively compare the proposed method with state-of-the-art point cloud renderers to show our advantages. Next, ablation studies are performed to validate the effect of each proposed module, including point-guided sampling, point encoder, and fusion decoder. Finally, we apply our method to point cloud applications.

\subsection{Experimental Settings}
\noindent\textbf{Datasets}. We perform experiments on indoor datasets containing point cloud and multi-view images, including ScanNet \cite{ScanNet} and ARKitScenes \cite{Arkitscenes}. 
ScanNet \cite{ScanNet} is an RGBD scanned dataset, which contains 2.5 million images at different views in 1,513 scenes. The dataset has been annotated with calibrated cameras and colored point clouds. We split the first 1,200 scenes as a training set and the rest as a testing set. ARkitScenes \cite{Arkitscenes} is a 3D indoor-scene understanding dataset, whose scenes are captured by Apple iPad Pro. There are around 5,000 scenes, and we choose the first 4,500 scenes for training and the 500 scenes for testing. 

\noindent\textbf{Metrics}. We adopt three common metrics, including PSNR, SSIM \cite{ssim}, and LPIPS \cite{lpips}, to evaluate the performance of Point2Pix.  They measure the reconstruction accuracy between the rendered images and ground-truth images.
%However, since point clouds are sparse, much information is missed. Reconstruction precision can not reflect the actual quality of rendered images. Thus, we also evaluate the performance by Frechet Inception Distance (FID) \cite{FID} metric which is frequently used to measure the distance between synthesized and real image sets.  

\noindent\textbf{Evaluation}.
We evaluate the rendering quality of different methods in two aspects, including \textit{non-finetuning} and \textit{finetuning}. The non-finetuning evaluation means directly measuring the rendering quality in the testing datasets. As for the finetuning evaluation, methods can refine their results on specific scenes to improve performance. Since finetuning evaluation in each case usually consumes much resources and time, we randomly choose 8 testing scenes from ScanNet datasets and the same number from the ArkitScene dataset for finetuning evaluation.

\noindent\textbf{Implementation Details}.
We adopt MinkUnet14A as our Point Encoder. The radius $r$ for point-guided sampling is $0.08$ meters. The maximal number $N$ of samples for each ray is 128. We extract feature volumes with $L=4$ scales. Thus the scales are  $\dfrac{1}{8}, \dfrac{1}{4}, \dfrac{1}{2}$, and $1$ respectively. The resolution of the final rendered images is $640 \times 480$.  During training, the initial learning rate is $0.004$ with AdamW \cite{AdamW} optimizer. It exponentially decays to $0.0004$ till the end (by 500 epochs). We set $\lambda_{pc}=0.1$, $\lambda_{nr}=1.0$, and  $\lambda_{per}=0.1$ empirically. The density threshold $D=10$. We train our model on 4 NVIDIA Titan-V GPUs, and the batch size is 1 for each GPU. 

\subsection{Comparison with Point Renderers}
We first compare our method with different point rendering methods by non-finetuning evaluation. After training, all methods are directly tested in novel scenes. The competitors include graphics-based point cloud renderer Pytorch3D \cite{Pytorch3d}, previous neural-based point renderers NPBG \cite{NPBG} and NPCR \cite{NPCR}, and image generator Pix2PixHD \cite{Pix2PixHD}. For Pix2PixHD, we use it as an image-to-image generator to translate the rendered images from graphics-based point renderer \cite{Pytorch3d} to the ground-truth images. The evaluation results are shown in Tab. \ref{tab:render_comparison}. Ours achieves significantly higher accuracy than other solutions, which reflects the great advantage in practical applications. 

\begin{figure*}[t]
	\centering
	\includegraphics[width=0.90\linewidth]{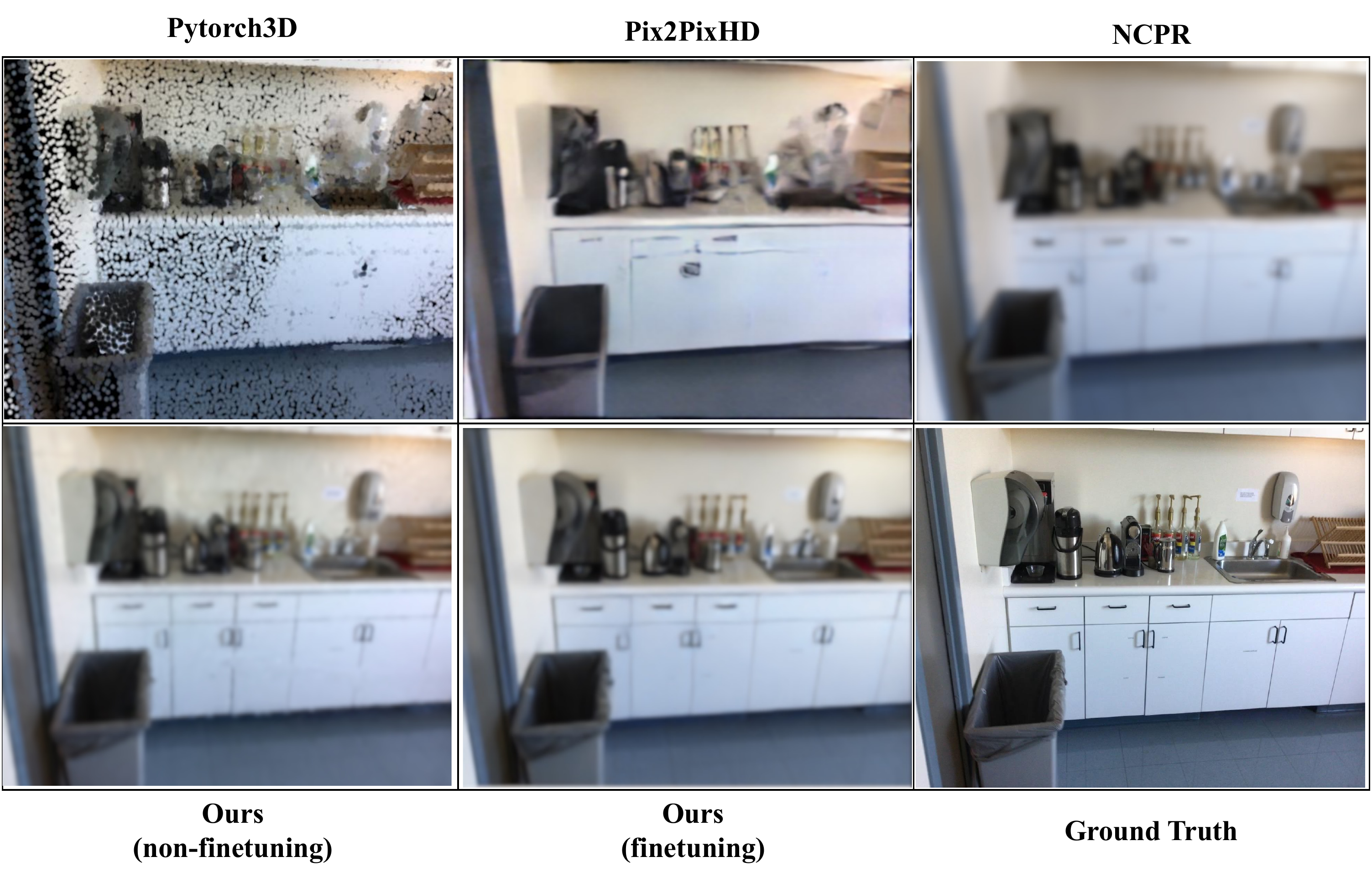}
	\vspace{-0.1 in}
	\caption{Qualitative comparison between different point renderers on the ScanNet \cite{ScanNet}.}
 \vspace{-0.1 in}
    \label{fig:scannet}
\end{figure*}

\subsection{Comparison with NeRF-based Synthesis}
We also compare the proposed method with NeRF-based synthesis in both non-finetuning and finetuning evaluations. In this experiment, we adopt the same coarse-to-fine ray sampling as NeRF-based methods \cite{NeRF,PlenOctree} for a fair comparison. The results are illustrated in Tab. \ref{tab:nerf_comparison}. 

To achieve general view synthesis in novel scenes, MVSNeRF \cite{MVSNeRF} and IBRNet \cite{IBRNet} combine image prior with NeRF, while ours combines the point cloud prior. For a fair comparison, we also pre-train these two methods on the same pretraining ScanNet dataset. Our method achieves the highest performance among all. Although NeRF \cite{NeRF}, NSVF \cite{NSVF}, and PlenOctrees \cite{PlenOctree} can achieve competitive accuracy, their training time is much longer. When training Instant-NGP \cite{Instant-NGP}, Plenoxels \cite{Plenoxels}, Point-NeRF \cite{PointNeRF}, and our Point2Pix in 20 minutes, ours achieves better performance. This experiment demonstrates the advantage of point cloud prior when combined with NeRF. 

\begin{table}[t]
\Huge
	\begin{adjustbox}{width=1.0\columnwidth,center}
		\begin{tabular}{c|c|ccc}
			\hline
			Method  &   \quad Time \quad &\quad PSNR($\uparrow$) \quad & \quad SSIM ($\uparrow$) & \quad LPIPS($\downarrow$) \quad \\ \hline
			% MVSNeRF \cite{MVSNeRF} & 0 mins & 17.11 & 0.771 & 0.356 \\ 
			% IBRNet \cite{IBRNet}& 0 mins & 17.49 & 0.784 & 0.339  \\ 
			Point-NeRF \cite{PointNeRF} & 0 mins & 17.53 & 0.685 & 0.517  \\ 
			\textbf{Point2Pix (Ours) } & 0 mins & \textbf{18.47} & \textbf{0.723} & \textbf{0.484}\\\hline
			NeRF \cite{NeRF} & $\sim$30 hours & 21.33 & 0.788 & 0.355 \\
			NSVF \cite{NSVF} & $\sim$40 hours & 22.47 & 0.791 & 0.337 \\
			PlenOctrees \cite{PlenOctree} & $\sim$30 hours & 22.02 & 0.795 & 0.341 \\
			Instant-NGP \cite{Instant-NGP} &20 mins & 21.94 & 0.775 & 0.363\\ 
			Plenoxels \cite{Plenoxels} & 20 mins & 22.35 & 0.780 & 0.346 \\ 
            Point-NeRF \cite{PointNeRF} & 20 mins & 22.55 & 0.792 & 0.336 \\
			\hline
			\textbf{Point2Pix (Ours)} & 20 mins & \textbf{23.02} & \textbf{0.815} & \textbf{0.318 }\\ \hline
		\end{tabular}
	\end{adjustbox}
	%%\vspace{-0.1in}
	\caption{Comparing our method with NeRF-based methods on the ScanNet dataset  \cite{ScanNet}.  ``Time" means the average finetuning time for all scenes. }
	%\vspace{-0.1in}
	\label{tab:nerf_comparison}
\end{table}

\subsection{Qualitative Comparison}
\begin{figure*}[t]
	\centering
	\includegraphics[width=0.90\linewidth]{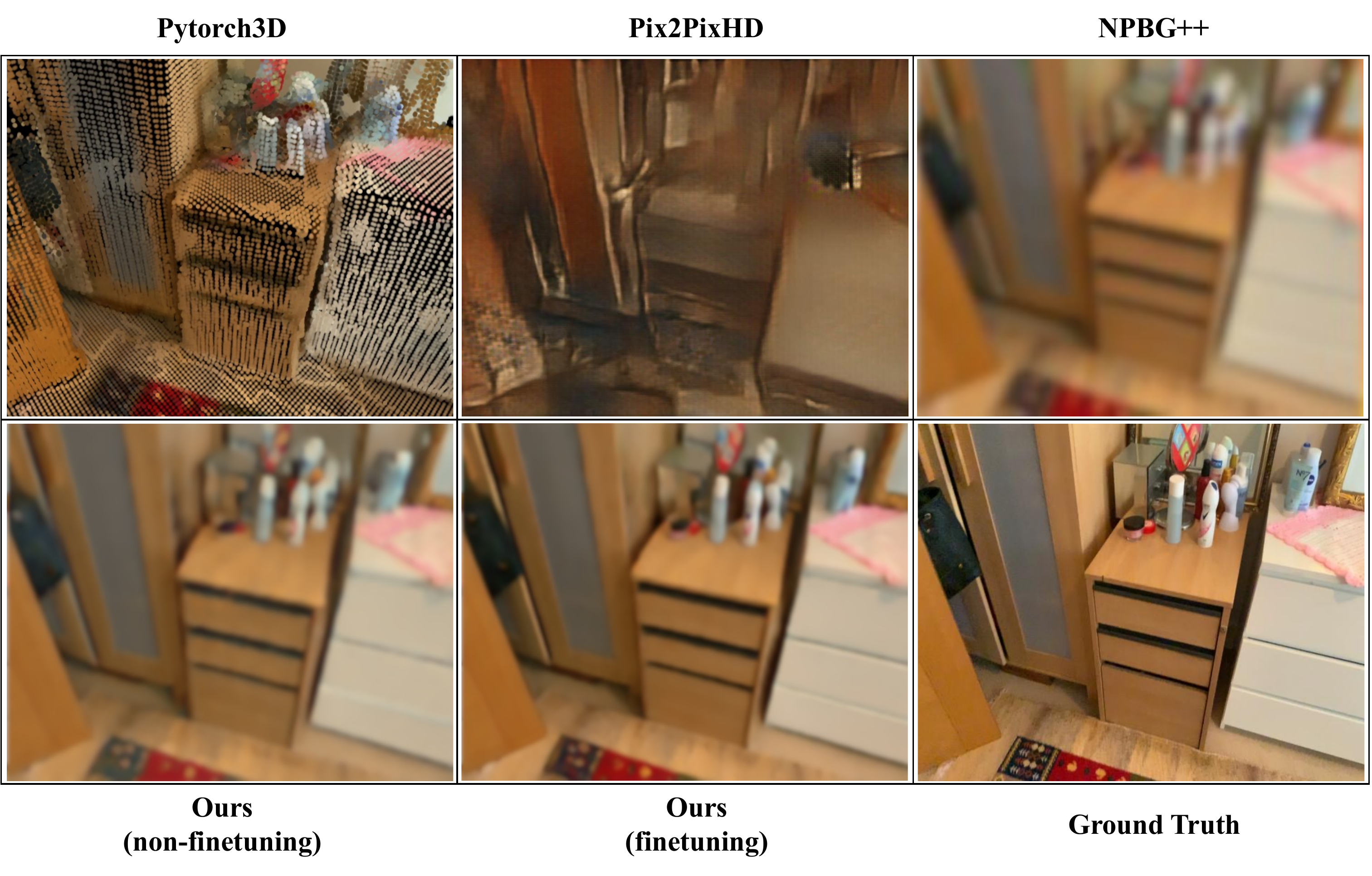}	\vspace{-0.1 in}
	\caption{Qualitative comparison between different point renderers and NeRF-based methods on the ArkitScenes \cite{Arkitscenes} dataset. }
	\label{fig:arkit}
\end{figure*}

We also qualitatively compare our Point2Pix with other point renderers and NeRF-based synthesis. The visualization is illustrated in Fig. \ref{fig:scannet} and Fig. \ref{fig:arkit}. The Graphics-based point renderer Pytorch3D \cite{Pytorch3d} usually generates images with holes because of sparse points. 
%%Because of missing 3D prior, the generated images are not realistic. 
Due to missing 3D prior, the generated images are not realistic. 
%%Ours achieves the best visual quality, which shows the superiority of our Point2Pix. 
Ours achieves the best visual quality, which shows Point2Pix's superiority.  

\subsection{Ablation Studies}
\noindent\textbf{Effect of Point-guided Sampling}.
To prove the efficiency of our point-guided sampling, we replace our sampling with other strategies used in NeRF-based methods, including uniform \cite{MVSNeRF,IBRNet} and coarse-to-fine sampling \cite{NeRF,PlenOctree}. Uniform sampling means uniformly obtaining $N$ points on each ray in near-to-far depth. Coarse-to-fine sampling is proposed by the original NeRF \cite{NeRF}. The coarse stage uniformly samples $\frac{N}{2}$ points and the fine stages samples $\frac{N}{2}$ points according to the probability from the coarse stage. 

Our point-guided sampling uniformly samples $N$ points while only inferring the valid samples. The results are shown in Tab. \ref{tab:ray_sampling}. 
Since our sampling strategy considers the point cloud 3D prior, we significantly decrease the average number of samples and reduce the rendering time and GPU memory. 

\begin{table}[t]
\Huge
	\begin{adjustbox}{width=1.0\columnwidth,center}
		\begin{tabular}{c|cccc}
			\hline
			%%%Ray Sampling & \quad \# Val. Sampl. \quad & \quad PSNR ($\uparrow$) \quad &  \quad \makecell{Render Time\\ (seconds, $\downarrow$)} \quad &  \quad \makecell{Train Memory \\ (GB, $\downarrow$)} \quad \\ \hline
			Ray Sampling & \# Val. Sampl. & PSNR ($\uparrow$) &  \makecell{Render Time\\ (seconds, $\downarrow$)}  &   \makecell{Train Memory \\ (GB, $\downarrow$)} \\ \hline
			Uniform  & 128 & 17.96 & 3.13 & 8.54 \\
			Coarse-to-Fine & 128 & \textbf{18.69}  & 5.78 & 9.17 \\
			Point-Guide (Ours) & \textbf{15.6} & \textbf{18.47} & \textbf{0.92} & \textbf{6.68} \\ \hline
		\end{tabular}
	\end{adjustbox}
	\vspace{-0.1in}
	\caption{Comparison between different ray sampling strategies on ScanNet \cite{ScanNet}. In our method, the mean sampling number for each ray is only $15.6$, which is significantly less than the others. }
	\vspace{-0.1in}
	\label{tab:ray_sampling}
\end{table}

\noindent\textbf{Effect of Multi-scale Radiance Fields}.
Previous methods, like GIRAFFE \cite{GIRAFFE}, HeadNeRF \cite{HeadNeRF}, StyleNeRF, and CIPS-3D \cite{CIPS-3D}, also render feature maps via NeRF, which can effectively reduce the memory and rendering time. However, different from ours, they only adopt a single scale of radiance fields. We validate our Multi-scale Radiance Fields in this experiment and show results in Tab. \ref{tab:multi_scale}. 

We first study the effect of the number of rays. In the condition of a single scale, if we directly render the image at the final resolution by NeRF, the memory consumption is heavy, and the running time is long. Decreasing sampled rays and increasing upsampling scale can promote the synthesis quality and rendering efficiency. With the increasing number of NeRF scales, accuracy further improves, which validates the effectiveness of our multi-scale radiance fields. We finally adopt the combination in the last column to balance the accuracy and time. 

\begin{table*}[t]
	\begin{adjustbox}{width=0.8\linewidth,center}
		\begin{tabular}{c|cc|ccc}
			\hline
			\# Scales &  \quad1 \quad \quad & 1 \quad & \quad 1  \quad &\quad 2 \quad&  \quad4 \quad \\ 
			\# Rays & $640 \times 480 $ & $ 320 \times 240$  & $ 80 \times 60 $ & $80 \times 60 $ &  $80 \times 60 $\\ 
			Upsampling & $ \times 1$ &  $ \times 2 $ & $ \times 8 $ & $ \times 8$ & $ \times 8 $ \\  \hline
			PSNR ($\uparrow$) & 17.49 & 17.86 & 18.05 & 18.16 & \textbf{18.47}  \\
			Rendering Time (seconds, $\downarrow$) & 13.12 & 3.56 & 0.85 & 0.92& 0.96   \\
			Training Memory (GB, $\downarrow$) & 22.93 & 11.12 & 6.27& 6.68 & 6.95   \\ 
            \hline
		\end{tabular}
	\end{adjustbox}	
	\vspace{-0.1 in}
				\caption{Effect of different combinations in terms of the number of scales, number of rays and scale of upsampling. We choose the combination in the last column, which achieves the best performance and is also efficient. }
	\label{tab:multi_scale}
\end{table*}

\noindent\textbf{Selection of Point Encoder}.
In our Point2Pix, the Point Encoder is the backbone to provide multi-scale 3D features. In the literature of point cloud analysis, many point-based networks \cite{PN,PointNet++,SparseConvNet,ME} were proposed. We compare different backbones in this experiment. 

The candidate networks include PointNet++ \cite{PointNet++}, SparseConvNet \cite{SparseConvNet}, and MinkUnet \cite{ME} (MinkUnet14A and MinkUnet34C). We evaluate them by only changing the point encoder, and the results are shown in Tab. \ref{tab:point_encoder}. PointNet++ \cite{PointNet++} consumes the largest memory and takes the longest rendering time, while the final synthesized accuracy is low. MinkUnet achieves the best results and is faster than SparseConvNet \cite{SparseConvNet}. We select MinkUnet14A as our point encoder since it is more efficient than MinkUnet34C. 

\begin{table}[t]
\Huge
	\begin{adjustbox}{width=1.0\columnwidth,center}
		\begin{tabular}{c|ccc}
			\hline
			%%Point Encoder & \quad \makecell{Training Memory \\ (GB, $\downarrow$)} \quad & \quad \makecell{Rendering Time \\ (seconds, $\downarrow$)} \quad & \quad \makecell{ PSNR \\ (dB, $\uparrow$) } \quad \\ \hline
			Point Encoder & \makecell{Training Memory \\ (GB, $\downarrow$)} &  \makecell{Rendering Time \\ (seconds, $\downarrow$)} & \makecell{ PSNR \\ (dB, $\uparrow$) }\\ \hline
			PointNet++ \cite{PointNet++} & 9.14 & 3.41 & 16.53 \\
			SparseConvNet \cite{SparseConvNet} & 8.18 & 2.18 & 18.26 \\
			MinkUnet14A \cite{ME} & \textbf{6.95} & \textbf{0.96} & \textbf{18.47} \\
			MinkUnet34C \cite{ME} & 8.26 & 1.45 & 18.45 \\ \hline
		\end{tabular}
	\end{adjustbox}
	\vspace{-0.1in}
    \caption{Comparison between different point encoders. We adopt MinkUnet14A  \cite{ME} to extract the basic 3D prior since it achieves accurate synthesized results and is also lightweight. }
    \vspace{-0.1in}
	\label{tab:point_encoder}
\end{table}

\noindent\textbf{Effect of Fusion Decoder}.
Previous neural point encoders usually adopt image-to-image translator \cite{UNet,Pix2PixHD} to render images from projected feature maps. We conduct this experiment to analyze the effect of our Fusion Decoder. We construct different alternatives by combining different decoder and fusion strategies, as shown in Tab. \ref{tab:fusion_decoder}. 

The combination between U-Net \cite{UNet} and concatenation strategy is the most frequently adopted \cite{NPBG,NPCR,ADOP}, while its performance is not high. When replacing the decoder with PixelShuffle \cite{PixelShuffle}, accuracy improves. It shows that the neural renderer does not require large respective fields as U-Net. When the concatenate strategy is replaced with our proposed fusion model, the performance is further improved, showing the rationality of our design. %We introduce the SPADE \cite{SPADE} (LayerNorm) module to fuse different feature maps. It achieves the best performance. 

\begin{table}[t]
\Huge
	\begin{adjustbox}{width=1.0\columnwidth,center}
		\begin{tabular}{c|c|ccc}
			\hline
			%%\quad Decoder \quad & \quad Fusion Strategy \quad &  \quad \makecell{PSNR \\ ($\uparrow$)} \quad & \quad \makecell{SSIM \\ ($\uparrow$)} \quad & \quad \makecell{LPIPS \\ ($\downarrow$)} \quad  \\ \hline
			Decoder & Fusion Strategy &  \makecell{PSNR \\ ($\uparrow$)} &  \makecell{SSIM \\ ($\uparrow$)}  & \makecell{LPIPS \\ ($\downarrow$)} \quad  \\ \hline
			U-Net \cite{UNet} & Concatenate & 18.04 & 0.708 & 0.511  \\ 
			PixelShuffle \cite{PixelShuffle} & Concatenate & 18.13 & 0.712 & 0.499  \\ 
			PixelShuffle \cite{PixelShuffle} & SPADE (LayerNorm) & \textbf{18.47} &\textbf{ 0.723} & \textbf{0.484}  \\ \hline
		\end{tabular}
	\end{adjustbox}
	\vspace{-0.1in}
	\caption{Comparison between different neural renderers. }
	\vspace{-0.1in}
	\label{tab:fusion_decoder}
\end{table}

\noindent\textbf{Effect of Point Cloud Loss}.
We perform this experiment to validate the effect of point cloud loss. By setting different loss weight $\lambda_{pc}$, we obtain the results in Tab. \ref{tab:pc_loss}. We conclude that point cloud loss indeed promotes the mapping process from point features to 3D attributes, thus improving the performance. Interestingly, larger $\lambda_{pc}$ does not necessarily achieve better accuracy, which demonstrates the minor difference between point cloud attributes with NeRF's attributes.

\begin{table}[t]
	\begin{adjustbox}{width=1.0\columnwidth,center}
	\begin{tabular}{c|p{1.8cm}<{\centering} p{1.8cm}<{\centering} p{1.8cm}<{\centering}}
		\hline
		$\lambda_{pc}$ & 0.0 & 0.1 & 1.0 \\ \hline
		PSNR (dB, $\uparrow$) & 18.23 & \textbf{18.47} & 18.30 \\
		\hline
	\end{tabular}
	\end{adjustbox}
	\vspace{-0.1in}
	\caption{The Effect of point cloud loss.}
	\label{tab:pc_loss}
\end{table}

\subsection{Application: Point Cloud Sampling}
\begin{figure}[t]
	\centering
	\includegraphics[width=1.0\linewidth]{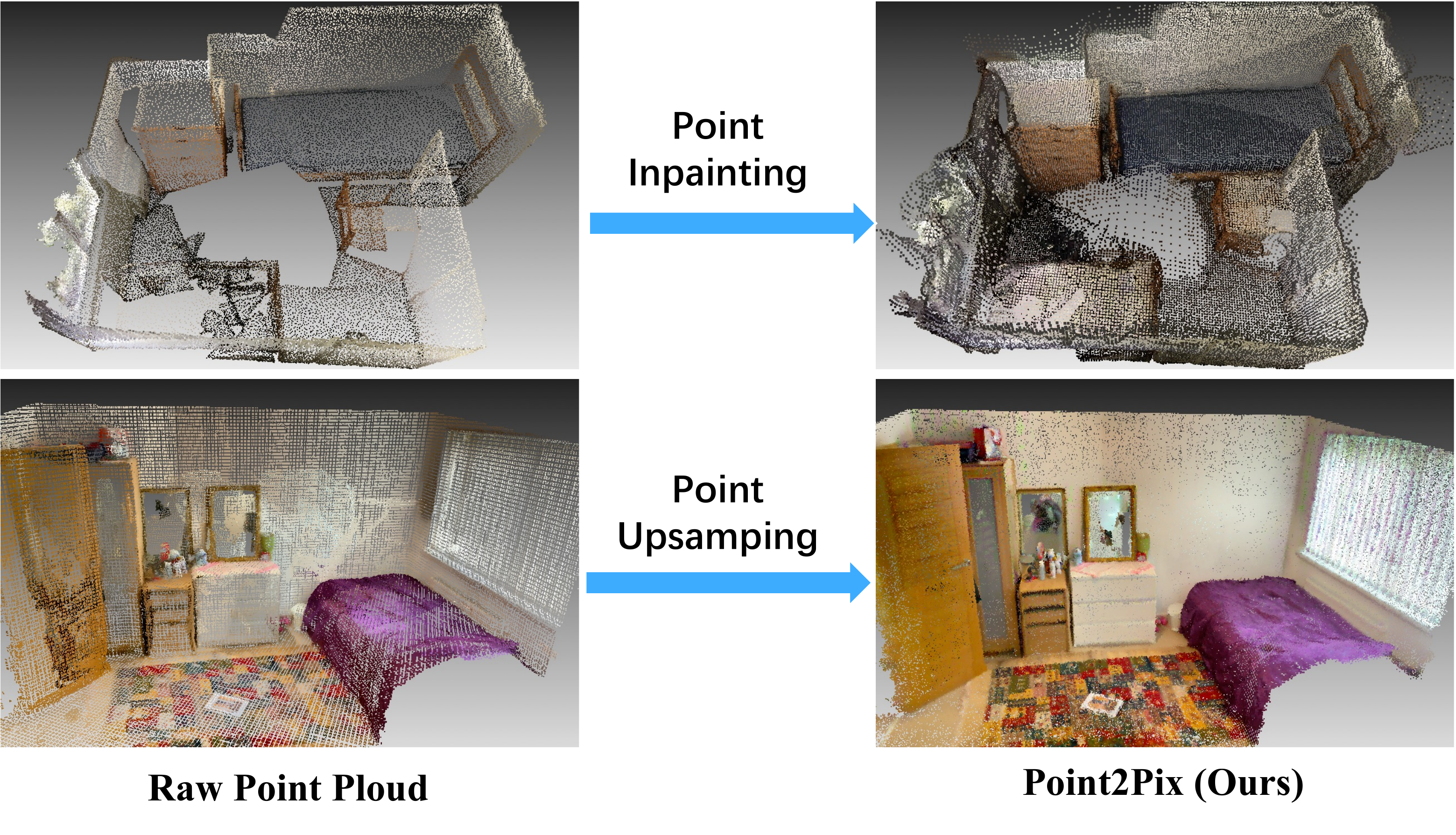}
	\caption{Our Point2Pix application in point cloud in-painting and upsampling. We upsample and fix the missing part of the raw point cloud by random and dense sampling around existing points.}
	\label{fig:upsampling}
	\vspace{-0.2 in}
\end{figure}

Since our point encoder can extract multi-scale point features and predict the density and color attributes, we upsample the raw point cloud by dense sampling and predict corresponding 3D attributes in the nearby area of existing points. As illustrated in Fig. \ref{fig:upsampling}, although there are no ground-truth dense points for us to perform supervised learning, our Point2Pix can still in-paint the missing points and insert many details for input point clouds. 

\section{Conclusion}
In this paper, we have proposed a general point renderer, which can be directly utilized to render photo-realistic images in indoor scenes. We introduce the advantages of point cloud representation to NeRF where existing points provide ground truth pairs during training, the point area can guide the ray sampling process and the 3D prior feature can be generalized to novel scenes. We propose Multi-scale Radiance Fields to extract discriminative 3D features, point-guided sampling to efficiently reduce the number of valid samples, and a Fusion Decoder to synthesize realistic images. Experiments and ablation studies demonstrate that our Point2Pix achieves state-of-the-art synthesized performance. Our Point2Pix can also be directly employed to upsample and in-paint the raw point cloud for indoor scenes. 

\noindent\textbf{Limitation and Future Work}. There are still common limitations in our proposed Point2Pix. First, the overall rendering time is long compared with recent caching-based rendering \cite{Plenoxels}. Second, apart from indoor scenes, it is still difficult to directly render photo-realistic images for arbitrary environments.  In future work, we will accelerate the rendering speed by combining it with other 3D scene representations, such as octrees. We will also extend our present work to more real-world situations, like the human body.
%or city street. 

\section*{Acknowledgments}
This work is partially supported by Shenzhen Science and Technology Program KQTD20210811090149095.

%%%%%%%%% REFERENCES
{\small
\bibliographystyle{ieee_fullname}
\bibliography{cvpr.bib}
}

\end{document}